%% file: paper.tex
\documentclass[runningheads]{llncs}\usepackage[]{graphicx}\usepackage[]{color}
\makeatletter
\def\maxwidth{ %
  \ifdim\Gin@nat@width>\linewidth
    \linewidth
  \else
    \Gin@nat@width
  \fi
}
\makeatother

\definecolor{fgcolor}{rgb}{0.345, 0.345, 0.345}

\usepackage{framed}
\makeatletter
 {\par\unskip\endMakeFramed%
 \at@end@of@kframe}
\makeatother

\definecolor{shadecolor}{rgb}{.97, .97, .97}
\definecolor{messagecolor}{rgb}{0, 0, 0}
\definecolor{warningcolor}{rgb}{1, 0, 1}
\definecolor{errorcolor}{rgb}{1, 0, 0}
\newenvironment{knitrout}{}{} 

\usepackage{alltt}
\usepackage{amsmath,amssymb,array}
\usepackage{xcolor}
\usepackage{graphicx}
\usepackage{xargs}
\usepackage{sparklines}
\renewcommand{\sparklineheight}{2}
\usepackage[utf8]{inputenc}
\usepackage{booktabs}
\usepackage{todonotes}
\usepackage[square,numbers,sort&compress,sectionbib]{natbib}
\renewcommand\citet{\cite}

\usepackage[hyphens]{url}
\usepackage{hyperref}

\usepackage{colortbl}
\usepackage{csquotes} 
\usepackage{mathtools}
\usepackage{float} 
\usepackage{adjustbox}
\usepackage{dsfont}
\usepackage[ruled, vlined, linesnumbered]{algorithm2e}
\DontPrintSemicolon                        
\SetKwInput{KwInput}{Input}                
\SetKwInput{KwOutput}{Output}              
\let\oldnl\nl
\newcommand{\nonl}{\renewcommand{\nl}{\let\nl\oldnl}}
\usepackage{cleveref}

\input{./latex-math/basic-ml.tex}
\input{./latex-math/basic-math.tex}

\newcounter{mycomment}

\newcommand{\xij}{x_{j}^{(i)}}                                         
\newcommand{\falej}{f_{j,ALE}}                                        
\newcommand{\faleme}{f_{ALE1st}}

\IfFileExists{upquote.sty}{\usepackage{upquote}}{}
\begin{document}

\title{Quantifying Model Complexity via Functional Decomposition for Better Post-Hoc Interpretability}
\titlerunning{Quantifying Model Complexity}

\author{
	Christoph Molnar 
	\and Giuseppe Casalicchio 
	\and Bernd Bischl 
}
\authorrunning{C. Molnar et al.}

\institute{Department of Statistics, LMU Munich, \\
  Ludwigstr. 33, 80539 Munich, Germany \\
  \email{christoph.molnar@stat.uni-muenchen.de}
}

\maketitle

\begin{abstract}
Post-hoc model-agnostic interpretation methods such as partial dependence plots can be employed to interpret complex machine learning models.
While these interpretation methods can be applied regardless of model complexity, they can produce misleading and verbose results if the model is too complex, especially w.r.t. feature interactions.
To quantify the complexity of arbitrary machine learning models, we propose model-agnostic complexity measures based on functional decomposition: number of features used, interaction strength and main effect complexity.
We show that post-hoc interpretation of models that minimize the three measures is more reliable and compact.
Furthermore, we demonstrate the application of these measures in a multi-objective optimization approach which simultaneously minimizes loss and complexity.
\end{abstract}

\section{Introduction}
\label{sec:introduction}

Machine learning models are optimized for predictive performance, but it is often required to understand models, e.g., to debug them, gain trust in the predictions, or satisfy regulatory requirements.
Many post-hoc interpretation methods either quantify effects of features on predictions, compute feature importances, or explain individual predictions, see \citep{molnar2019,guidotti2018survey} for more comprehensive overviews.
While model-agnostic post-hoc interpretation methods can be applied regardless of model complexity \citep{ribeiro2016model}, their reliability and compactness deteriorates when models use a high number of features, have strong feature interactions and complex feature main effects.
Therefore, model complexity and interpretability are deeply intertwined and reducing complexity can help to make model interpretation more reliable and compact.
Model-agnostic complexity measures are needed to strike a balance between interpretability and predictive performance \citep{ruping2006learning,bibal2016interpretability}.

\textbf{Contributions.}
We propose and implement three model-agnostic measures of machine learning model complexity which are related to post-hoc interpretability.
To our best knowledge, these are the first model-agnostic measures that describe the global interaction strength, complexity of main effects and number of features.
We apply the measures to different datasets and machine learning models.
We argue that minimizing these three measures improves the reliability and compactness of post-hoc interpretation methods.
Finally, we illustrate the use of our proposed measures in multi-objective optimization.

\section{Related Work and Background}
\label{sec:related}

In this section, we introduce the notation, review related work, and describe the functional decomposition on which we base the proposed complexity measures.

\textbf{Notation:} We consider machine learning prediction functions $f:\mathbb{R}^p \mapsto \mathbb{R}$, where $f(x)$ is a prediction (e.g., regression output or a classification score).
For the decomposition of $f$, we write $f_S:\mathbb{R}^{|S|} \mapsto \mathbb{R}$, $S \subseteq\{1, \ldots, p\}$, to denote a function that maps a vector $x_S \in \mathbb{R}^{|S|}$ with a subset of features to a marginal prediction.
If subset $S$ contains a single feature $j$, we write $f_j$.
We refer to the training data of the machine learning model with the tuples $\D = \{(x^{(i)},y^{(i)})\}_{i=1}^n$ and refer to the value of the $j$-th feature from the $i$-th instance as $x_j^{(i)}$.
We write $X_j$ to refer to the $j$-th feature as a random variable.

\textbf{Complexity and Interpretability Measures:}
\label{sec:other}
In the literature, model complexity and (lack of) model interpretability are often equated.
Many complexity measures are model-specific, i.e., only models of the same class can be compared (e.g., decision trees).
Model size is often used as a measure for interpretability (e.g., number of decision rules, tree depth, number of non-zero coefficients) \citep{huysmans2011empirical,ruping2006learning,askira1998knowledge,yang2017scalable,schielzeth2010simple,lakkaraju2017interpretable,furnkranz2012foundations,ustun2016supersparse}.
Akaikes Information Criterion (AIC) 
and the Bayesian Information Criterion (BIC)
are more widely applicable measures for the trade-off between goodness of fit and degrees of freedom.
In \citep{philipp2018measuring}, the authors propose model-agnostic measures of model stability.
In \citep{plumb2019regularizing}, the authors propose explanation fidelity and stability of local explanation models.
Further approaches measure interpretability based on experimental studies with humans, e.g., whether humans can predict the outcome of the model \citep{zhou2018measuring,huysmans2011empirical,dhurandhar2017tip,poursabzi2018manipulating,friedler2019assessing}.

\textbf{Functional Decomposition:}
\label{sec:decomposition}
Any high-dimensional prediction function can be decomposed into a sum of components with increasing dimensionality:

\begin{eqnarray}\label{eqn:decomp} f(x)  = &\overbrace{f_0}^\text{Intercept} + \overbrace{\sum_{j=1}^p f_j(x_j)}^\text{1st order effects} + \overbrace{\sum_{j<k}^p f_{jk}(x_j, x_k)}^\text{2nd order effects} + \ldots + \overbrace{f_{1,\ldots,p}(x_1, \ldots, x_p)}^\text{p-th order effect}
\end{eqnarray}
This decomposition is only unique with additional constraints regarding the components.
Accumulated Local Effects (ALE) were proposed in \cite{apley2016visualizing} as a tool for visualizing feature effects (e.g., Figure~\ref{fig:c-demo}) and as unique decomposition of the prediction function with components $f_S = f_{S,ALE}$.
The ALE decomposition is unique under an orthogonality-like property described in \citep{apley2016visualizing}.


The ALE main effect $f_{j,ALE}$ of a feature $x_j, j \in \{1,\ldots,p\}$ for a prediction function $f$ is defined as
\begin{eqnarray}\label{eqn:ale}
\falej(x_j) = \int_{z_{0,j}}^{x_j} \mathbb{E}\left[\frac{\partial f(X_1,\ldots,X_p)}{\partial X_j}\middle|X_j = z_j\right]dz_j-c_j
\end{eqnarray}
Here, $z_{0,j}$ is a lower bound of $X_j$ (usually the minimum of $x_j$) and the expectation $\mathbb{E}$ is computed conditional on the value for $x_j$ and over the marginal distribution of all other features.
The constant $c_j$ is chosen so that the mean of $f_{j,ALE}(x_j)$ with respect to the marginal distribution of $X_j$ is zero, so that the ALE components sum to the full prediction function.
By integrating the expected derivative of $f$ with respect to $X_j$ the effect of $x_j$ on the prediction function $f$ is isolated from the effects of all other features.
ALE main effects are estimated with finite differences, i.e., access to the gradient of a prediction function is not required (see \citep{apley2016visualizing}).
We base our proposed measures on the ALE decomposition, because ALE are computationally cheap (worst case $O(n)$ per main effect), they can be computed sequentially instead of simultaneously, they do not require knowledge of the joint distribution, and several software implementations exist \citep{iml,alepackage}.

\section{Functional Complexity}
\label{sec:measures}

In this section, we motivate complexity measures based on functional decomposition.
Based on Equation~\ref{eqn:decomp}, we decompose the prediction function into a constant (estimated as $f_0 = \frac{1}{n}\sum_{i=1}^n f(\xi)$), main effects (estimated by ALE), and a remainder term containing interactions (i.e., the difference between the full model and constant + main effects).

\begin{eqnarray} f(x) 
=&\underbrace{f_0 + \sum_{j=1}^p \overbrace{\falej(x_j)}^\text{MEC: How complex?} + \overbrace{IA(x)}^{\text{IAS: Interaction strength?}}}_{\text{NF: How many features were used?}}
\end{eqnarray}
This arrangement of components emphasizes a decomposition of the prediction function into a main effect model and an interaction remainder.
We can analyze how well the main effect model itself approximates $f$ by looking at the magnitude of the interaction measure IAS.
The average main effect complexity (MEC) captures how many parameters are needed to describe the one-dimensional main effects on average.
The number of features used (NF) describes how many features were used in the full prediction function.

\subsection{Number of Features (NF)}
\label{sec:nfeatures}

We propose an approach based on feature permutation to determine how many features are used by a model.
We regard features as "used" when changing a feature changes the prediction.
If available, the model-specific number of features is preferable.
The model-agnostic version is useful when the prediction function is only accessible via API or when the machine learning pipeline is complex.


The proposed procedure is formally described in Algorithm~\ref{algo:nfeat}.
To estimate whether the $j$-th feature was used, we sample instances from data $\D$, replace their $j$-th feature values with random values from the distribution of $X_j$ (e.g., by sampling $\xj$ from other instances from $\D$), and observe whether the predictions change.
If the prediction of any sample changes, the feature was used. 
\begin{algorithm}
\caption{Number of Features Used (NF)}\label{algo:nfeat}
\KwInput{Number of samples $M$, data $\D$}
NF = 0\;
	\For{$j \in 1,\ldots,p$}{
		Draw $M$ instances $\{x^{(m)}\}_{m=1}^M$ from dataset $\D$\;
			Create $\{x^{(m)*}\}_{m=1}^M$ as a copy of $\{x^{(m)}\}_{m=1}^M$ \;
			\For{$m\in 1,\ldots,M$}{
				Sample $\xj^{(new)}$ from $\{\xij\}_{i=1}^n$ with the constraint that $\xj^{(new)} \neq \xj^{(m)}$\;
				Set $\xj^{(m)*} = \xj^{(new)}$\;
			}
			\lIf{$\fh(x^{(m)*}) \neq \fh(x^{(m)}) \text{ for any }  m \in \{1,\ldots,M\}$}{$NF = NF + 1$.
		}
		}
\Return NF
\end{algorithm}

We tested the NF heuristic with the Boston Housing data.
We trained decision trees (CART) with maximum depths $\in\{1,2,10\}$ leading to 1, 2 and  4 features used and an L1-regularized linear model with penalty $\lambda\in\{10, 5, 2, 1, 0.1, 0.001\}$ leading to 0, 2, 3, 4, 11 and 13 features used.
For each model, we estimated NF with sample sizes $M\in\{10, 50, 500\}$ and repeated each estimation 100 times.
For the elastic net models, NF was always equal to the number of non-zero weights. 
For CART, the mean absolute differences between NF and number of features used in the trees were 0.280 ($M=10$), 0.020 ($M=50$) and 0.000 ($M=500$).

\subsection{Interaction Strength (IAS)}
\label{sec:interaction}

Interactions between features mean that the prediction cannot be expressed as a sum of independent feature effects, but the effect of a feature depends on values of other features \citep{molnar2019}.
We propose to measure interaction strength as the scaled approximation error between the ALE main effect model and the prediction function $f$.
Based on the ALE decomposition, the ALE main effect model is defined as the sum of first order ALE effects:
$$\faleme(x) = f_0 + f_{1,ALE}(x_1) + \ldots + f_{p,ALE}(x_p)$$
We define interaction strength as the approximation error measured with loss $L$:
\begin{eqnarray}\label{eqn:pre}
IAS = \frac{\mathbb{E}(L(f, \faleme))}{\mathbb{E}(L(f, f_0))} \geq 0
\end{eqnarray}
Here, $f_0$ is the mean of the predictions and can be interpreted as the functional decomposition where all feature effects are set to zero.
IAS with the $L2$ loss equals 1 minus the R-squared measure, where the true targets $y_i$ are replaced with $f(\xi)$.
$$IAS = \frac{\sum_{i=1}^n(f(\xi)-\faleme(\xi))^2}{\sum_{i=1}^n(f(\xi) - f_0)^2} = 1 - R^2$$
If $IAS=0$, then $L(f,\faleme)=0$, which means that the first order ALE model perfectly approximates $f$ and the model has no interactions.

\subsection{Main Effect Complexity (MEC)}
\label{sec:curve}


To determine the average shape complexity of ALE main effects $\falej$, we propose the main effect complexity (MEC) measure.
For a single ALE main effect, we define $\text{MEC}_j$ as the number of parameters needed to approximate the curve with piece-wise linear models.
For the entire model, MEC is the average $\text{MEC}_j$ over all main effects, weighted with their variance.
Figure~\ref{fig:c-demo} shows an ALE plot (= main effect) and its approximation with two linear segments.
\begin{knitrout}\small
\definecolor{shadecolor}{rgb}{0.969, 0.969, 0.969}\color{fgcolor}\begin{figure}

{\centering \includegraphics[width=10cm,height=4cm]{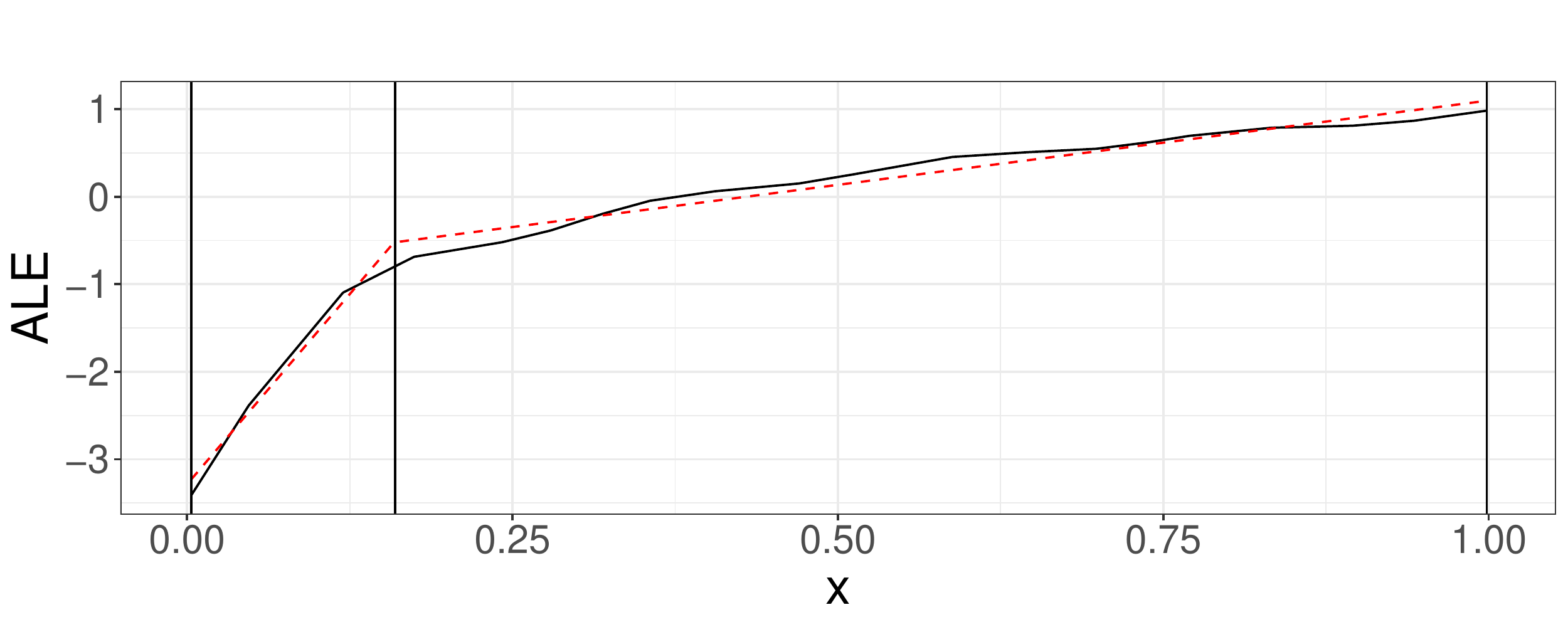} 

}

\caption[ALE curve (solid line) approximated by two linear segments (dotted line)]{ALE curve (solid line) approximated by two linear segments (dotted line).}\label{fig:c-demo}
\end{figure}

\end{knitrout}
We use piece-wise linear regression to approximate the ALE curve.
Within the segments, linear models are estimated with ordinary least squares.
The breakpoints that define the segments are found by greedy and exhaustive search along the interval boundaries of the ALE curve.
Greedy here means that we first optimize the first breakpoint, then the second breakpoint with the first breakpoint fixed and so on.
We measure the degrees of freedom as the number of non-zero coefficients for intercepts and slopes of the linear models.
The approximation allows some error, e.g., an almost linear main effect may have $\text{MEC}_j=1$, even if dozens of parameters would be needed to describe it perfectly. 
The approximation quality is measured with R-squared ($R^2$), i.e., the proportion of variance of $\falej$ that is explained by the approximation with linear segments.
An approximation has to reach an $R^2 \geq 1-\epsilon$, where $\epsilon$ is the user defined maximum approximation error.
We also introduced parameter $max_{seg}$, the maximum number of segments.
In the case that an approximation cannot reach an $R^2 \geq 1-\epsilon$ with a given $max_{seg}$, $\text{MEC}_j$ is computed with the maximum number of segments.
The selected maximum approximation error $\epsilon$ should be small, but not too small.
We found $\epsilon$ between $0.01$ and $0.1$ visually meaningful (i.e. a subjectively good approximation) and used $\epsilon=0.05$ throughout the paper.
We apply a post-processing step that greedily sets slopes of the linear segments to zero, as long as $R^2\in\{1-\epsilon,1\}$.
The post-processing potentially decreases the $\text{MEC}_j$, especially for models with constant segments like decision trees.
$\text{MEC}_j$ is averaged over all features to obtain the global main effect complexity.
Each $\text{MEC}_j$ is weighted with the variance of the corresponding ALE main effect to give more weight to features that contribute more to the prediction.
Algorithm~\ref{algo:AMEC} describes the MEC computation in detail.

\begin{algorithm}
\caption{Main Effect Complexity (MEC).}\label{algo:AMEC}
	\KwInput{Model $f$, approximation error $\epsilon$, max. segments $max_{seg}$, data $\D$}
	Define $R^2(g_j, \falej) := \sum_{i=1}^n (g_j(\xij) - \falej(\xij))^2 / \sum_{i=1}^n (\falej(\xij))^2$\;
	\For{$j \in \pset$}{
        Estimate $\falej$\;
	\tcp{Approximate ALE with linear model}
	Fit $g_j(x_j) = \beta_0 + \beta_1 x_j$ predicting $\falej(\xij)$ from $\xij$, $i\in{1,\ldots,n}$\;
	Set $K=1$\;
	\tcp{Increase nr. of segments until approximation is good enough}
	\While{$K < max_{seg}$ AND $R^2(g_j,\falej) < (1- \epsilon)$}{
	        \tcp{Find intervals $Z_k$ through exhaustive search along ALE curve breakpoints}
		\tcp{For categorical feature, set slopes $\beta_{1,k}$ to zero}
		$g_j(x_j) = \sum_{k=1}^{K+1} \I_{\xj\in Z_k}\cdot\left(\beta_{0,k} + \beta_{1,k}\xj\right)$\;
		Set $K = K+1$
	}
	Greedily set slopes to zero while $R^2>1-\epsilon$\;
\tcp{Sum of non-zero coefficients minus first intercept}
	$MEC_j = K + \sum_{k=1}^K \mathbb{I}_{\beta_{1,k} > 0} - 1$\;
$V_j = \frac{1}{n}\sum_{i=1}^n (\falej(x^{(i)}))^2$\;
}
	\Return{$MEC = \frac{1}{\sum_{j=1}^pV_j}\sum_{j=1}^p V_j \cdot MEC_j$\;}
\end{algorithm}

\section{Application of Complexity Measures}
\label{sec:experiment}

In the following experiment, we train various machine learning models on different prediction tasks and compute the model complexities.
The goal is to analyze how the complexity measures behave across different datasets and models.
The dataset are:
Bike Rentals \citep{bike} (n=731; 3 numerical, 6 categorical features),
Boston Housing (n=506; 12 numerical, 1 categorical features),
(downsampled) Superconductivity \citep{hamidieh2018data} (n=2000; 81 numerical, 0 categorical features) and
Abalone \citep{uci} (n=4177; 7 numerical, 1 categorical features).

\begin{table}[ht]
\centering
\begingroup\fontsize{8pt}{9pt}\selectfont
\begin{tabular}{r|rrrr|rrrr|rrrr|rrrr|}
  \hline
  & \multicolumn{4}{c|}{bike}& \multicolumn{4}{c|}{Boston Housing}& \multicolumn{4}{c|}{superconductivity}& \multicolumn{4}{c|}{abalone}\\ learner& MSE& MEC& IAS& NF& MSE& MEC& IAS& NF& MSE& MEC& IAS& NF& MSE& MEC& IAS& NF\\ \hline
cart & 923035 & 1.1 & 0.07 & 6 & 23.7 & 1.9 & 0.12 & 4 & 325.0 & 1.0 & 0.23 & 8 & 6.0 & 2.8 & 0.09 & 3 \\ 
  cart2 & 1245105 & 1.0 & 0.01 & 2 & 29.8 & 1.7 & 0.02 & 2 & 417.6 & 1.0 & 0.22 & 3 & 6.7 & 3.0 & 0.02 & 1 \\ 
  cvglmnet & 667291 & 1.1 & 0.00 & 9 & 27.4 & 1.0 & 0.00 & 8 & 351.1 & 1.0 & 0.00 & 50 & 5.1 & 1.0 & 0.00 & 6 \\ 
  gamboost & 539538 & 1.6 & 0.00 & 8 & 17.7 & 2.5 & 0.00 & 10 & 360.3 & 1.7 & 0.00 & 14 & 5.3 & 1.1 & 0.00 & 4 \\ 
  ksvm & 424184 & 1.6 & 0.04 & 8 & 13.7 & 1.7 & 0.09 & 13 & 256.0 & 2.2 & 0.25 & 81 & 4.6 & 1.0 & 0.12 & 8 \\ 
  lm & 629144 & 1.5 & 0.00 & 9 & 23.4 & 1.0 & 0.00 & 13 & 337.4 & 1.0 & 0.00 & 81 & 4.9 & 1.0 & 0.00 & 8 \\ 
  rf & 478115 & 1.8 & 0.06 & 9 & 13.2 & 2.5 & 0.10 & 13 & 167.4 & 3.0 & 0.25 & 81 & 4.6 & 1.7 & 0.30 & 8 \\ 
   \hline
\end{tabular}
\endgroup
\caption{Model performance and complexity on 4 regression tasks for various learners: linear models (lm), cross-validated regularized linear models (cvglmnet), kernel support vector machine (ksvm), random forest (rf), gradient boosted generalized additive model (gamboost), decision tree (cart) and decision tree with depth 2 (cart2).} 
\label{tab:exp}
\end{table}

Table \ref{tab:exp} shows performance and complexity of the models.
As desired, the main effect complexity for linear models is 1 (except when categorical features with 2+ categories are present as in the bike data), and higher for more flexible methods like random forests.
The interaction strength (IAS) is zero for additive models (boosted GAM, (regularized) linear models).
Across datasets we observe that the underlying complexity measured as the range of MEC and IAS across the models varies.
The bike dataset seems to be adequately described by only additive effects, since even random forests, which often model strong interactions show low interaction strength here.
In contrast, the superconductivity dataset is better explained by models with more interactions.
For the abalone dataset there are two models with low MSE: the support vector machine and the random forest.
We might prefer the SVM, since main effects can be described with single numbers ($MEC = 1$) and interaction strength is low.


\section{Improving Post-hoc Interpretation}
\label{sec:post-hoc}

Minimizing the number of features (NF), the interaction strength (IAS), and the main effect complexity (MEC) improves reliability and compactness of post-hoc interpretation methods such as partial dependence plots, ALE plots, feature importance, interaction effects and local surrogate models.

\textbf{Fewer features, more compact interpretations.}
Minimizing the number of features improves the readability of post-hoc analysis results.
The computational complexity and output size of most interpretation methods scales with $O(\text{NF})$, like feature effect plots \citep{apley2016visualizing,friedman2001greedy} or feature importance \citep{fisher2018all,casalicchio2018visualizing}.
As demonstrated in Table~\ref{tab:spark-table-multiobj}, a model with fewer features has a more compact representation.
If additionally $IAS=0$, the ALE main effects fully characterize the prediction function.
Interpretation methods that analyze 2-way feature interactions scale with $O(\text{NF}^2)$.
A complete functional decomposition requires to estimate $\sum_{k=1}^{NF} {NF\choose k}$ components which has a computational complexity of $O(2^{NF})$.

\textbf{Less interaction, more reliable feature effects.}
Feature effect plots such as partial dependence plots and ALE plots visualize the marginal relationship between a feature and the prediction.
The estimated effects are averages across instances.
The effects can vary greatly for individual instances and even have opposite directions when the model includes feature interactions.

In the following simulation, we trained three models with different capabilities of modeling interactions between features: a linear regression model, a support vector machine (radial basis kernel, C=0.05), and gradient boosted trees.
We simulated 500 data points with 4 features and a continuous target based on \citep{friedman1991multivariate}.
Figure \ref{fig:pdp-unreliable} shows an increasing interaction strength depending on the model used.
More interaction means that the feature effect curves become a less reliable summary of the model behavior.

\begin{knitrout}\small
\definecolor{shadecolor}{rgb}{0.969, 0.969, 0.969}\color{fgcolor}\begin{figure}

{\centering \includegraphics[width=12cm,height=4.5cm]{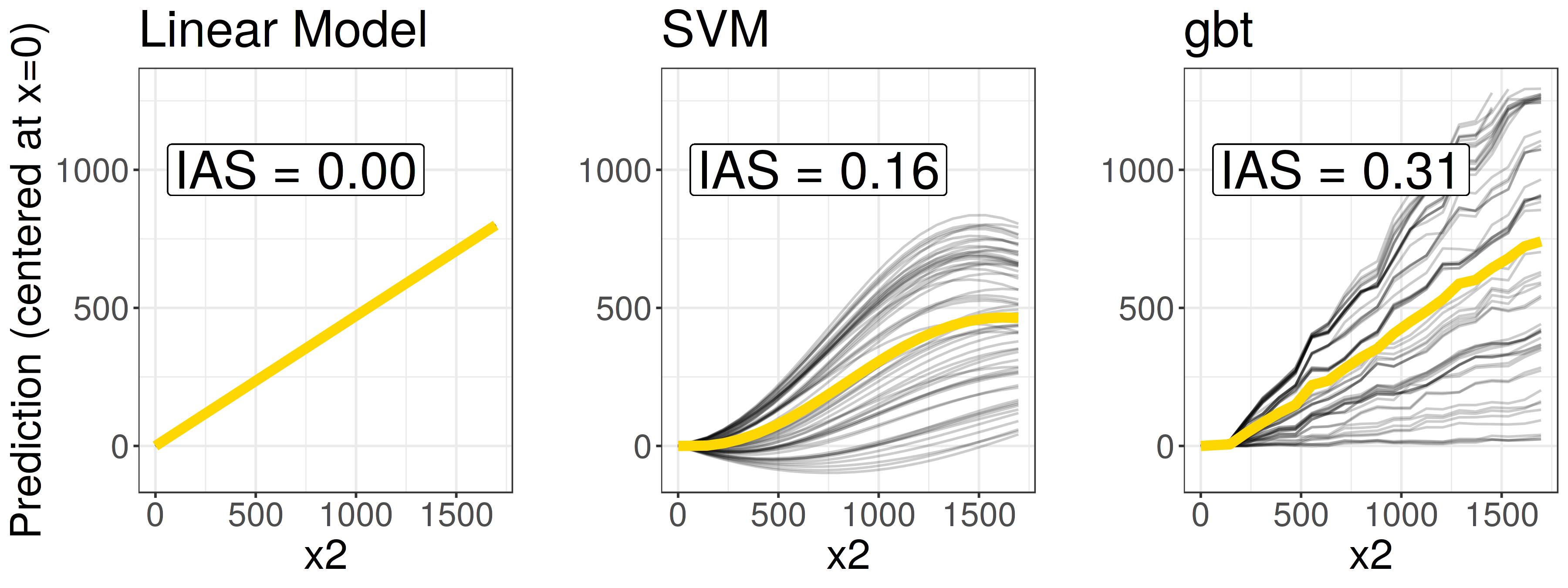} 

}

\caption[The higher the interaction strength in a model (IAS increases from left to right), the less representative the Partial Dependence Plot (light thick line) becomes for individual instances represented by their Individual Conditional Expectation curves (dark thin lines)]{The higher the interaction strength in a model (IAS increases from left to right), the less representative the Partial Dependence Plot (light thick line) becomes for individual instances represented by their Individual Conditional Expectation curves (dark thin lines).}\label{fig:pdp-unreliable}
\end{figure}

\end{knitrout}

\textbf{The less complex the main effects, the better summarizable.}
In linear models, a feature effect can be expressed by a single number, the regression coefficient.
If effects are non-linear the method of choice is visualization  \citep{apley2016visualizing,friedman2001greedy}.
Summarizing the effects with a single number (e.g., using average marginal effects \citep{leeper2017interpreting}) can be misleading, e.g., the average effect might be zero for U-shaped feature effects.
As a by-product of MEC, there is a third option: Instead of reporting a single number, the coefficients of the segmented linear model can be reported.
Minimizing MEC means preferring models with main effects that can be described with fewer coefficients, offering a more compact model description.

\section{Application: Multi-objective Optimization}
\label{sec:multiobj}

%
%

%
%
%

%

%
%

We demonstrate model selection for performance and complexity in a multi-objective optimization approach.
For this example, we predict wine quality (scale from 0 to 10) \citep{cortez2009modeling} from the wines physical-chemical properties such as alcohol and residual sugar of 4870 white wines.
It is difficult to know the desired compromise between model complexity and performance before modeling the data.
A solution is multi-objective optimization \citep{freitas2014comprehensible}.
We suggest searching over a wide spectrum of model classes and hyperparameter settings, which allows to select a suitable compromise between model complexity and performance.

We used the mlrMBO model-based optimization framework \citep{horn2016multi} with ParEGO \citep{knowles2006parego} (500 iterations) to find the best models based on four objectives: number of features used (NF), main effect complexity (MEC), interaction strength (IAS) and cross-validated mean absolute error (MAE) (5-fold cross-validated).
We optimized over the space of following model classes (and hyperparameters): \textbf{CART} (maximum tree-depth and complexity parameter cp), \textbf{s}upport \textbf{v}ector \textbf{m}achine (cost C and inverse kernel width sigma), \textbf{elastic net} regression (regularization alpha and penalization lambda), \textbf{g}radient \textbf{b}oosted \textbf{t}rees (maximum depth, number of iterations), gradient \textbf{boost}ed \textbf{g}eneralized \textbf{a}dditive \textbf{m}odel (number of iterations nrounds) and \textbf{r}andom \textbf{f}orest (number of split features mtry).

\textbf{Results}.
The multi-objective optimization resulted in 27 models.
The measures had the following ranges: MAE 0.41 -- 0.63, number of features 1 --  11, mean effect complexity 1 -- 9 and interaction strength 0 -- 0.71.
For a more informative visualization, we propose to visualize the main effects together with the measures in Table~\ref{tab:spark-table-multiobj}.
The selected models show different trade-offs between the measures.

\begin{table}[ht]
\centering
\caption{A selection of four models from the Pareto optimal set, along with their ALE main effect curves. From left to right, the columns show models with 1) lowest MAE, 2) lowest MAE when $MEC=1$, 3) lowest MAE when $IAS =\leq 0.2$, and 4) lowest MAE with $NF \leq 7$.} 
\label{tab:spark-table-multiobj}
\begin{tabular}{l|p{2.2cm}p{2.2cm}p{2.2cm}p{2.2cm}}
  \hline
 & gbt (maxdepth:8, nrounds:269) & svm (C:23.6979, sigma:0.0003) & gbt (maxdepth:3, nrounds:98) & CART (maxdepth:14, cp:0.0074) \\ 
  \hline
MAE & 0.41 & 0.58 & 0.52 & 0.59 \\ 
  MEC & 4.2 & 1 & 4.5 & 2 \\ 
  IAS & 0.64 & 0 & 0.2 & 0.2 \\ 
  NF & 11 & 11 & 11 & 4 \\
\hline
\end{tabular}
\end{table}

\section{Discussion}
\label{sec:discussion}

We proposed three measures for machine learning model complexity based on functional decomposition: number of features used, interaction strength and main effect complexity.
Due to their model-agnostic nature, the measures allow model selection and comparison across different types of models and they can be used as objectives in automated machine learning frameworks.
This also includes "white-box" models:
For example, the interaction strength of interaction terms in a linear model or the complexity of smooth effects in generalized additive models can be quantified and compared across models.
We argued that minimizing these measures for a machine learning model improves its post-hoc interpretation.
We demonstrated that the measures can be optimized directly with multi-objective optimization to make the trade-off between performance and post-hoc interpretability explicit.

\textbf{Limitations.}
The proposed decomposition of the prediction function and definition of the complexity measures will not be appropriate in every situation.
For example, all higher order effects are combined into a single interaction strength measure that does not distinguish between two-way interactions and higher order interactions.
However, the framework of accumulated local effect decomposition allows to estimate higher order effects and to construct different interaction measures.
The main effect complexity measure only considers linear segments but not, e.g., seasonal components or other structures.
Furthermore, the complexity measures quantify machine learning models from a functional point of view and ignore the structure of the model (e.g., whether it can be represented by a tree).
For example, main effect complexity and interaction strength measures can be large for short decision trees (e.g. in Table \ref{tab:exp}). 

\textbf{Implementation.}
The code for this paper is available at \url{https://github.com/compstat-lmu/paper_2019_iml_measures}.
For the examples and experiments we relied on the mlr package \citep{JMLR:v17:15-066} in R \citep{r2018}.

\textbf{Acknowledgements.}
This work is funded by the Bavarian State Ministry of Science and the Arts in the framework of the Centre Digitisation.Bavaria (ZD.B) and supported by the German Federal Ministry of Education and Research (BMBF) under Grant No. 01IS18036A.
The authors of this work take full responsibilities for its content.

\bibliographystyle{splncs04}
\bibliography{Bib}

\end{document}

%% file: latex-math/basic-ml.tex
\newcommand{\pset}{\{1, \ldots, p\}}                                        
\newcommand{\D}{\mathcal{D}}                                                
\renewcommand{\xi}[1][i]{x^{(#1)}}                                          
\newcommand{\xj}{x_j}                                                       

\newcommand{\fh}{f}                                                   

%% file: latex-math/basic-math.tex
\ifdefined\N                                                                
\renewcommand{\N}{\mathds{N}}                                                
\else
  \newcommand{\N}{\mathds{N}}
\fi
\ifdefined\C
  \renewcommand{\C}{\mathds{C}}                                             
\else
  \newcommand{\C}{\mathds{C}}
\fi

\newcommand{\I}{\mathbb{I}}                                                 

